\documentclass{article}
\usepackage{spconf,amsmath,graphicx}
\usepackage{amsthm}
\newtheorem{theorem}{Theorem} 

\usepackage{amsmath}
\usepackage{dsfont}
\usepackage{algorithm}
\usepackage{mathtools}
\usepackage{algpseudocode}
\usepackage{algpseudocode}
\usepackage{amsfonts}
\usepackage[bottom]{footmisc}
\newcommand\scalemath[2]{\scalebox{#1}{\mbox{\ensuremath{\displaystyle #2}}}}

\title{A New Probabilistic Distance Metric with Application in Gaussian Mixture Reduction}
\name{Ahmad Sajedi$^{1, 2}$, Yuri A. Lawryshyn$^{2}$, Konstantinos N. Plataniotis$^{1}$}
\address{$^{1}$ The Edward S. Rogers Sr. Department of Electrical \& Computer Engineering, University of Toronto\\
$^{2}$ Centre for Management of Technology \& Entrepreneurship (CMTE), University of Toronto}
\begin{document}
%
\maketitle
\begin{abstract}
This paper presents a new distance metric to compare two continuous probability density functions. The main advantage of this metric is that, unlike other statistical measurements, it can provide an analytic, closed-form expression for a mixture of Gaussian distributions while satisfying all metric properties. These characteristics enable fast, stable, and efficient calculations, which are highly desirable in real-world signal processing applications. The application in mind is Gaussian Mixture Reduction (GMR), which is widely used in density estimation, recursive tracking, and belief propagation. To address this problem, we developed a novel algorithm dubbed the Optimization-based Greedy GMR (OGGMR), which employs our metric as a criterion to approximate a high-order Gaussian mixture with a lower order. Experimental results show that the OGGMR algorithm is significantly faster and more efficient than state-of-the-art GMR algorithms while retaining the geometric shape of the original mixture.
\end{abstract}
\section{Introduction}
\label{sec:intro}
Multidimensional statistical elements are usually described by their probabilistic descriptors in the form of continuous probability density functions (PDFs) in a well-defined space, such as the Hilbert space of density functions \cite{egozcue2006hilbert, principe2010information}. 
Probabilistic descriptors are rarely available or difficult to analyze, but they can be approximated to any degree of conciseness. A common way to estimate an unknown continuous PDF is by using a linear expansion of well-known and simple probability distributions derived from partitioning theory. Researchers often use Gaussian distribution as a basis function and approximate the descriptor with a mixture of Gaussian (MoG) \cite{ardeshiri2012mixture, d2021likeness, kampa2011closed, runnalls2007kullback, chen2010constraint, assa2018wasserstein}. In fact, MoG is a modular architecture that estimates probabilistic descriptors from a weighted combination of normal distributions. As a formal expression, let ${x} \in \mathbb{R}^{n}$ be a vector variable in a $n$-dimensional space, i.e., ${x} = \{x_{1}, x_{2}, \cdots, x_{n}\}$. We then define the MOG as the weighted combination of $M$ Gaussian distributions:
\vspace{-5pt}
\begin{flalign} \label{eq1}
P({x}) = \sum_{m=1}^{M} \pi_{m} \mathcal{N}({x}|{\mu_{m}}, {\Sigma_{m}}),
\end{flalign}
\setlength{\parindent}{0pt}where, $\sum_{m=1}^{M}\pi_{m} = 1$, $\pi_{m} \geq 0, \forall m \in\{1, \cdots, M\}$, and a component $\mathcal{N}({x}|{\mu_{m}}, {\Sigma_{m}})$ is a normal distribution with a mean vector ${\mu_{m}} \in \mathbb{R}^{n}$ and a covariance matrix ${\Sigma_{m}} \in \mathbb{S}^{n}_{++}$.

\setlength{\parindent}{0.25in}In addition to approximating the probabilistic descriptors, the efficiency of the similarity search should also be considered. Signal processing and machine learning algorithms typically use one of several dissimilarity measurements to compare two PDFs \cite{kullback1951information, runnalls2007kullback, kampa2011closed, hoang2015cauchy, bhattacharyya1943measure, pandy2022transferability, ruschendorf1985wasserstein, zhu2022gaussian}. When PDF descriptors are not available, we can compare the underlying MoG approximations using various dissimilarities such as Kullback-Leibler divergence \cite{runnalls2007kullback, ardeshiri2012mixture}, Cauchy-Schwartz divergence \cite{kampa2011closed, nielsen2012closed}, Bhattacharyya dissimilarity \cite{mohammadi2015improper}, Wasserstein distance \cite{assa2018wasserstein}, and Likeness-based dissimilarities \cite{d2021likeness, d2022fixed, principe2010information, asad2021propel}. Some of these measures fail to meet boundness and/or metric properties \cite{runnalls2007kullback, kampa2011closed, d2021likeness, d2022fixed, principe2010information}, while others are incapable of providing analytical closed-form expressions for MoGs \cite{runnalls2007kullback, ardeshiri2012mixture, mohammadi2015improper, assa2018wasserstein}. A closed-form expression for distances allows for fast and efficient computation, which is vital for applications involving high-dimensional data. Taking all of the above into account, our research aim is to define a distance measure that satisfies all metric properties, particularly triangle inequality, while providing a closed-form solution to MoGs without imposing any constraints, which is an open research problem.

\setlength{\parindent}{0.25in}The proposed metric is a foundational solution to many signal processing and machine learning problems. Our main focus will be on an application area where the number of components in MoG grows exponentially, like multi-target tracking \cite{ardeshiri2012mixture} and nonlinear filtering \cite{tam1999adaptive}. In such scenarios, Gaussian mixture reduction (GMR) is necessary to control the growth of a mixture. To that end, we will use the proposed metric as a criterion to approximate a high-order MoG with a lower order using a new algorithm called Optimization-based Greedy GMR (OGGMR). The results show that our algorithm outperforms the state-of-the-art while being computationally more efficient (see Fig.\ref{fig1}). The contributions of this paper are:
\vspace{-0.12cm}
\begin{itemize}
    \item We propose a true distance that follows the metric properties while providing closed-form solutions to the distance between MoGs. It has benefits in computations and in many analyses relying on metric properties.
    \vspace{-0.17cm}
    \item We design an OGGMR algorithm using the proposed distance to efficiently solve the GMR problem. 
    
    \vspace{-0.17cm}
    \item We conduct an experimental analysis on a GMR scenario to demonstrate the effectiveness and efficiency of the proposed distance and the OGGMR algorithm.
    \vspace{-0.14cm}
\end{itemize}

\section{Methodology} \label{sec:ProposedMetric}
\subsection{Proposed Probabilistic Distance Metric} \label{subsec2-1}
In this section, we build a true metric probability distance that yields a closed-form expression for real-valued Gaussian and MoG distributions, whose solutions depend on the number of basic components, distribution parameters (mean and covariance), and mixture coefficients. The following theorem proposes a bounded metric in the probability space that, unlike prior measures, captures magnitude and angular distances.

\begin{theorem} \label{th1}
Let $p({x})$ and $q({x})$ be continuous PDFs on a probability space $\Omega$. Take both to be Hilbert space $L^{2}$ elements of square-integrable functions, i.e., $\int_{\Omega}p({x})^{2}d{x}$ and $\int_{\Omega}q({x})^{2}d{x}$ are well characterized. Then
\begin{flalign} 
\scalemath{0.97}{
d_{NCP}(p, q) = \sqrt{K_{NCP}(p, p)+K_{NCP}(q, q) - 2K_{NCP}(p, q)}} \nonumber
\end{flalign}
\vspace{-20pt}
\begin{flalign} \label{eq2}
 & \: \: \:\scalemath{0.97}{= \sqrt{2-2\frac{\int_{\Omega}p({x})q({x})d{x}}{\sqrt{\int_{\Omega}p({x})^{2}d{x}}\sqrt{\int_{\Omega}q({x})^{2}d{x}}}}}
\end{flalign}
is a true bounded distance in the space of probability distribution $\mathcal{P}$ and consequently $(\mathcal{P}, d_{NCP})$ is a metric space. The $K_{NCP}(p, q) = \frac{\int_{\Omega}p({x})q({x})d{x}}{\sqrt{\int_{\Omega}p({x})^{2}d{x}}\sqrt{\int_{\Omega}q({x})^{2}d{x}}}$ denotes the Normalized Cross-information Potential \cite{principe2010information} between $p$ and $q$. 
\end{theorem}
\begin{proof}
The $d_{NCP}(p, q)$ is a true metric since it fulfills all four properties of a metric for $\forall p, q$, and $r \in {\mathcal{P}}$, i.e.,\\
\textbf{Non-negativity: } $d_{NCP}(p,q) \geq 0$. Careful examination of the definition of $d_{NCP}$ and the limited range of $K_{NCP}$ ($0 \leq K_{NCP} \leq 1$) reveals that $0 \leq d_{NCP} \leq \sqrt{2}$.\\
\textbf{Identity of indiscernibles: } $d_{NCP}(p, q) = 0 \Longleftrightarrow p = q$.\\
$\scalemath{1}{\bullet \:\: d_{NCP} (p, q) = 0 \Longrightarrow K_{NCP}(p, q) = 1 \Longrightarrow p =kq, k\in{\mathbb{R}}}$\\  $\xRightarrow[]{\text{probability axiom}} p = q,$\\
$\scalemath{1}{\bullet \:\: p = q \Longrightarrow K_{NCP}(p, q) = 1 \Longrightarrow d_{NCP}(p, q) = 0}$. \\
\textbf{Symmetry: } Obviously, $d_{NCP}(p, q) = d_{NCP}(q, p)$.\\
\textbf{Triangle inequality: }$d_{NCP}(p, q) \leq d_{NCP}(p, r) + d_{NCP}(r, q)$. Due to the fact that $K_{NCP}(p, q)$ is a Mercer kernel, $d_{NCP}(p, q)$ can be described as the $l_{2}$ norm in the Hilbert space of square-integrable functions, i.e., $\big\|\phi(p) - \phi(q) \big\|_{L^{2}(\Omega)}$, where $\phi(\cdot)$ is the non-linear mapping from $\mathcal{P}$ to ${L^{2}(\Omega)}$. Therefore, the triangle inequality can be proven with ease. 
\begin{flalign}
   & d_{NCP}(p, q)  = \Big\|\phi(p) - \phi(q) \Big\|_{L^{2}(\Omega)} \leq \Big\|\phi(p) - \phi(r) \Big\|_{L^{2}(\Omega)}   \nonumber \\
    &  + \Big\|\phi(r) - \phi(q) \Big\|_{L^{2}(\Omega)} = d_{NCP}(p, r) + d_{NCP}(r, q). \nonumber \\&\text{This completes the proof of Theorem \ref{th1}. \:\:\:\:\:\:\:\:\:\:\:\:\:\:\:\:\:\:\:\:\:\:\:\:\:\:\:\:\qedhere  \nonumber} 
\end{flalign}
\end{proof}


This metric is indeed the combined angular and magnitude distances resulting from the definition of $K_{NCP}$ and the Euclidean distance ($l_{2}$ norm) in Hilbert space, respectively. In other words, our chordal-shape metric has the ability to measure the geometric differences between two arbitrary probability distributions. In addition, triangle inequality can help find the shortest path between two distributions, which is critical for embedding, clustering, and mixture approximations.
To provide a better perspective on the proposed metric distance, we derived the closed-form expression when the probability set is defined in the family of multivariate normal, or MoG, distributions. The closed-form solution can reduce the amount of effort needed to obtain a valid approximation by avoiding simulation methods such as Monte Carlo, which may significantly increase computation time and lead to a loss of precision \cite{kampa2011closed, d2022fixed, nielsen2012closed}. The following theorem sheds light on that and provides an analytical, closed-form expression for MoGs to facilitate the computation.
\begin{theorem} \label{th2}
Let $P({x}) = \sum_{m=1}^{M}\pi_{m}\mathcal{N}({x}|{{\mu_{p_{m}}}}, {{\Sigma_{p_{m}}}}) = \sum_{m=1}^{M}\pi_{m}p_{m}(x)$ and $Q({x}) = \sum_{n=1}^{N}\tau_{n}\mathcal{N}({x}|{{\mu_{q_{n}}}}, {{\Sigma_{q_{n}}}}) = \sum_{n=1}^{N}\tau_{n}q_{n}(x)$ be two finite MoGs. The proposed distance between $P({x})$ and $Q({x})$ can be expressed in closed form as:
\begin{flalign} \label{eq3}
d_{NCP}(P, Q) = \sqrt{2-2\frac{\boldsymbol{\pi}^{T}\boldsymbol{\Psi}_{PQ}\boldsymbol{\tau}}{\sqrt{\boldsymbol{\pi}^{T}\boldsymbol{\Psi}_{P}\boldsymbol{\pi}}\sqrt{\boldsymbol{\tau}^{T}\boldsymbol{\Psi}_{Q}\boldsymbol{\tau}}}},
\end{flalign}
where, $\boldsymbol{\pi} = [\pi_{1}, \cdots, \pi_{M}]^{T} \in (\mathbb{R}^{+})^{M}$ and $\boldsymbol{\tau} = [\tau_{1}, \cdots, \tau_{N}]^{T}$ $\in (\mathbb{R}^{+})^{N}$. For $i, j = 1, \cdots, M$ and $k, l = 1, \cdots, N$, the components of $\boldsymbol{\Psi}_{PQ}$, $\boldsymbol{\Psi}_{P}$, and $\boldsymbol{\Psi}_{Q}$ are computed as
\begin{flalign}
&\scalemath{0.911}{[\boldsymbol{\Psi}_{PQ}]_{ik} = \int_{\mathbb{R}^{n}}\mathcal{N}({x}|{\mu_{p_{i}}}, {\Sigma_{p_{i}}})\mathcal{N}({x}|{\mu_{q_{k}}}, {\Sigma_{q_{k}}})d{x} = (2\pi)^{-\frac{n}{2}}}  \nonumber\\
&\scalemath{0.911} {|{\Sigma_{p_{i}}}+{\Sigma_{q_{k}}}|^{-\frac{1}{2}}\exp(-\frac{1}{2}({\mu_{p_{i}}}-{\mu_{q_{k}}})^{T}({\Sigma_{p_{i}}}+{\Sigma_{q_{k}}})^{-1} ({\mu_{p_{i}}}-{\mu_{q_{k}}}))}, \nonumber \\[8pt]
&\scalemath{0.911}{[\boldsymbol{\Psi}_{P}]_{ij} = \int_{\mathbb{R}^{n}}\mathcal{N}({x}|{\mu_{p_{i}}}, {\Sigma_{p_{i}}})\mathcal{N}({x}|{\mu_{p_{j}}}, {\Sigma_{p_{j}}})d{x} =(2\pi)^{-\frac{n}{2}}}  \nonumber\\
&\scalemath{0.911}{|{\Sigma_{p_{i}}}+{\Sigma_{p_{j}}} |^{-\frac{1}{2}}\exp(-\frac{1}{2}({\mu_{p_{i}}}-{\mu_{p_{j}}})^{T}({\Sigma_{p_{i}}}+{\Sigma_{p_{j}}})^{-1}({\mu_{p_{i}}
}-{\mu_{p_{j}}}))}, \nonumber\\[8pt]
&\scalemath{0.911}{[\boldsymbol{\Psi}_{Q}]_{kl} = \int_{\mathbb{R}^{n}}\mathcal{N}({x}|{\mu_{q_{k}}}, {\Sigma_{q_{k}}})\mathcal{N}({x}|{\mu_{q_{l}}}, {\Sigma_{q_{l}}})d{x} = (2\pi)^{-\frac{n}{2}}} \nonumber\\
& \scalemath{0.911}{|{\Sigma_{q_{k}}}+{\Sigma_{q_{l}}} |^{-\frac{1}{2}}\exp(-\frac{1}{2}\big({\mu_{q_{k}}}-{\mu_{q_{l}}})^{T}({\Sigma_{q_{k}}}+{\Sigma_{q_{l}}} )^{-1}({\mu_{q_{k}}
}-{\mu_{q_{l}}})).} \nonumber 
\end{flalign}
\end{theorem}
\begin{proof}
The cross-information potential between MoGs is
\begin{flalign}
&\scalemath{0.811}{\int_{\mathbb{R}^{n}}P({x})Q({x})d{x} = \int_{\mathbb{R}^{n}} \sum_{m=1}^{M}\pi_{m}\mathcal{N}({x}|{{\mu_{p_{m}}}}, {{\Sigma_{p_{m}}}})\sum_{n=1}^{N}\tau_{n}\mathcal{N}({x}|{{\mu_{q_{n}}}}, {{\Sigma_{q_{n}}}})d{x}} \nonumber \\
&\scalemath{0.811}{= \sum_{m=1}^{M}\sum_{n=1}^{N} \pi_{m}\tau_{n} \int_{\mathbb{R}^{n}}\mathcal{N}({x}|{\mu_{p_{m}}}, {\Sigma_{p_{m}}})\mathcal{N}({x}|{\mu_{q_{n}}}, {\Sigma_{q_{n}}})d{x} \overset{(a)}{=} \boldsymbol{\pi}^{T}\boldsymbol{\Psi}_{PQ}\boldsymbol{\tau}},  \nonumber
\end{flalign}
where (a) is derived by integrating the products of two multivariate normal distributions, as shown in \cite{hogg2020data}.
Applying the same trick to self-information potentials, i.e., $\int_{\mathbb{R}^{n}}P(x)^{2}dx$ and $\int_{\mathbb{R}^{n}}Q(x)^{2}dx$ and then combining with Theorem \ref{th1} complete the proof of Theorem \ref{th2}. \:\:\:\:\:\:\:\:\:\:\:\:\:\:\:\:\:\qedhere
\end{proof}
It is worth noting that for low-dimensional Gaussian components (e.g., $n = 2, 3$ in tracking applications), Eq. $3$ has a computational complexity of $O(M^{2})$ to calculate cross-information potentials. If $n \gg M$ and $N$, the complexity will be $O(n^{3})$ due to covariance matrix inversion. However, both cases are still much lower than the Monte-Carlo method, which depends on the number of samples $S \gg M, N,$ and $n$. In the following subsection, we will demonstrate the efficacy of our metric by solving the GMR problem, which is widely used in signal processing.
\vspace{-5pt}
\subsection{Case Study: Gaussian Mixture Reduction}
\label{subsec2.2}
Gaussian mixtures are parametric tools for approximating probabilities due to their unique properties, including the ability to model multi-modalities and density functions. However, approximation accuracy always trades off with computational efficiency. Furthermore, in some applications, like multi-target tracking \cite{ardeshiri2012mixture}, and nonlinear filtering \cite{tam1999adaptive} the number of components grows exponentially over time. If no action is taken, the problem may become computationally intractable. Despite the fact that an MoG can have many components, its shape is typically simple, and we can produce an MoG with fewer components by combining local information and removing redundant components. For example, a single Gaussian may represent a mode of true density previously represented by multiple Gaussian components. This is known as the Gaussian Mixture Reduction (GMR) problem.

\vspace{-3pt}
\begin{algorithm} 
\small
\caption{Optimization-based Greedy GMR (OGGMR)}
\label{alg1}
\begin{algorithmic}
\State $1)$ (Greedy Initialization) Get the initial estimate of the mixture coefficients, the mean vector, and the covariance matrices of the reduced mixture using Algorithm \ref{alg2}. 
\State $2)$ (Refinement) Given the initial estimate, perform iterative optimization techniques such as SLSQP over the optimization problem to refine the estimates.
\end{algorithmic}
\end{algorithm}

\vspace{-5pt}
The GMR problem is defined as finding a Gaussian mixture $Q(x|\boldsymbol{\mu_{Q}}, \boldsymbol{\Sigma_{Q}}, \boldsymbol{\tau}) = \sum_{n=1}^{N}\tau_{n}\mathcal{N}(x|\mu_{q_{n}}, \Sigma_{q_{n}})$ that minimizes the dissimilarity $d(\cdot, \cdot)$ from the original Gaussian mixture, $P(x|\boldsymbol{\mu_{P}}, \boldsymbol{\Sigma_{P}}, \boldsymbol{\pi}) = \sum_{m=1}^{M}\pi_{m}\mathcal{N}(x|\mu_{p_{m}}, \Sigma_{p_{m}})$, where $N \leq M$. Formally, the solution can be found by solving the following optimization problem:
\begin{algorithm} 
\small
\caption{Greedy Reduction Algorithm}
\label{alg2}
\textbf{Input:} \text{Original MoG $P(x)$, Number of reduced components $N$}
\begin{algorithmic}[1]
\While{$M > N$}
\For{ $i = 1 : M$}\vspace{+0.05cm}
\State $\tilde{{\pi}}_{i} =  \dfrac{{\pi}_{i}}{\sqrt{|\Sigma_{p_{i}}|}}$
\EndFor
\State $i^{*} = \arg \min \tilde{{\pi}}_{i}$
\For{$j = 1 : M \setminus\{i^{*}\}$}\vspace{+0.05cm}
\State $S_{i^{*}j} = \dfrac{\int p_{i^{*}}(x)p_{j}(x)d{x}}{\sqrt{\int p_{i^{*}}(x)^{2}d{x}}\sqrt{\int p_{j}(x)^{2}d{x}}}$
\EndFor
\State $j^{*} = \arg \max S_{i^{*}j}$ \vspace{+0.05cm}
\State $\mu_{i^{*}} = \dfrac{\pi_{i^{*}}}{\pi_{i^{*}} + \pi_{j^{*}}}\mu_{p_{i^{*}}} + \dfrac{\pi_{j^{*}}}{\pi_{i^{*}} + \pi_{j^{*}}}\mu_{p_{j^{*}}}$
\State $\Sigma_{i^{*}} = \dfrac{\pi_{i^{*}}}{\pi_{i^{*}} + \pi_{j^{*}}}\Sigma_{p_{i^{*}}} + \dfrac{\pi_{j^{*}}}{\pi_{i^{*}} + \pi_{j^{*}}}\Sigma_{p_{j^{*}}}+\dfrac{\pi_{i^{*}}\pi_{j^{*}}}{(\pi_{i^{*}} + \pi_{j^{*}})^{2}}$ 
\Statex $\:\:\:\:\:\:\:\:(\mu_{p_{i^{*}}} - \mu_{p_{j^{*}}})(\mu_{p_{i^{*}}} - \mu_{p_{j^{*}}})^{T}$ 
\State $q_{i^{*}}(x) = \mathcal{N}(x|\mu_{i^{*}}, \Sigma_{i^{*}})$
\State $P(x) = P(x) - \pi_{i^{*}} p_{i^{*}}(x) - \pi_{j^{*}} p_{j^{*}}(x) + (\pi_{i^{*}}+\pi_{j^{*}}) q_{i^{*}}(x)$
\State $M = M - 1$
\EndWhile
\end{algorithmic}
\textbf{Output:} \text{Initial estimates of the reduced MoG's parameters}
\end{algorithm}


\vspace*{-0.48cm}
\begin{flalign} 
\min_{\boldsymbol{\mu_{Q}}, \boldsymbol{\Sigma_{Q}}, \boldsymbol{\tau}} \quad & d\Big(P\big(x|\boldsymbol{\mu_{P}}, \boldsymbol{\Sigma_{P}}, \boldsymbol{\pi}\big), Q\big(x|\boldsymbol{\mu_{Q}}, \boldsymbol{\Sigma_{Q}}, \boldsymbol{\tau}\big)\Big) \label{eq4}\\
\textrm{s.t.} \quad & {1}^{T}\boldsymbol{\tau} = 1 \nonumber\\
  & \boldsymbol{\tau} \succeq {0}  \nonumber  \\
  & {\Sigma_{q_{n}}} \succeq 0 \:\:\:\:\:\:\:\:\:\:\:\:\:\:\:\:\:\:\:\:\:\:\:\:\:\:\: n = 1, 2, \cdots, N \nonumber\\
  & {\Sigma_{q_{n}}} - {\Sigma_{q_{n}}}^{T} = 0 \:\:\:\:\:\:\:\:\:\:\:\:n = 1, 2, \cdots, N \nonumber.
\end{flalign}

\vspace*{-0.04cm}
This is a complex, nonconvex, nonlinear, constrained optimization problem whose optimization variables are the parameters of the reduced MoG. In order to retain the majority of information from the original MoG, different dissimilarity measures have been employed in the literature for mixture reduction \cite {mohammadi2015improper, assa2018wasserstein, crouse2011look, liu2012shape}. Despite this, some issues arise when tackling the GMR. First of all, none of these dissimilarities meet the triangle inequality to allow the shortest path between two MoGs. Another common problem among all dissimilarities is the presence of a large number of local minima. To address these issues, we propose an algorithm dubbed Optimization-based Greedy GMR (OGGMR) that employs greedy reduction along with refinement steps based on the proposed distance (Alg.\ref{alg1}). The coarse greedy reduction step is used to obtain a starting point for the refinement step in the OGGMR algorithm. Specifically, Alg. \ref{alg2} selects a Gaussian component with the lowest normalized weight based on the determinant of the covariance as a candidate (lines 2–5) and chooses another component closest to the candidate based on the proposed distance (lines 6–9). The selected candidate and its closest component are then merged into a single Gaussian using a moment-preserving merge to preserve the overall mean and covariance of the mixture, and the algorithm proceeds with the reduction (lines 10–15). After determining the initial estimate, we use numerical methods like SLSQP to solve the optimization problem (Eq. \ref{eq4}) in the refinement step while considering our distance as a measurement.
\vspace{-5.5pt}
\section{Experiments and Results}
\label{sec3}
\vspace{-5.5pt}
In this section, we describe the experimental setups that will be used throughout the study. Our goal is to use the proposed distance and the OGGMR algorithm for GMR problems.

\begin{table*}[htb]
\centering
\caption{The performance of the GMR algorithms in three reduction scenarios with different dissimilarity measurements.}
\scalebox{0.85}{
\begin{tabular}{c c c c c c c}
\hline
{\textbf{Algorithm}} & {\textbf{KL}} & {\textbf{ISE}} & {\textbf{NISE}} & {\textbf{TSL}} & {\textbf{CS}} & {\textbf{OURS}}\\
\hline
{West $(N=1)$} & $0.42587$ & $0.11084$ & $0.11237$ & $0.06466$ & $0.11479$ & $0.46572$\\
{Enhanced West $(N=1)$} & $0.42554$ & $0.11084$ & $0.11237$ & $0.06466$ & $0.11479$ & $0.46572$\\
{GMRC $(N=1)$} & $0.42589$ & $0.11084$ & $0.11237$ & $0.06466$ & $0.11479$ & $0.46572$\\
{OGGMR $(N=1)$} & $0.42564$ & $0.11084$ & $0.11237$ & $0.06466$ & $0.11479$ & $0.46572$\\
\hline
{West $(N=3)$} & $0.17090$ & $0.10518$ & $0.04586$ & $0.02576$ & $0.05129$ & $0.31661$\\
{Enhanced West $(N=3)$} & $0.21632$ & $0.08184$ & $0.05033$ & $0.03259$ & $0.05142$ & $0.46572$\\
{GMRC $(N=3)$} & $0.07509$ & $\mathbf{0.03632}$ & $\mathbf{0.03898}$ & $\mathbf{0.02179}$ & $0.04202$ & $0.27600$\\
{OGGMR  $(N=3)$} & $\mathbf{0.07257}$ & $0.03983$ & $\mathbf{0.03898}$ & $0.02400$ & $\mathbf{0.02393}$ & $\mathbf{0.21747}$\\
\hline
{West $(N=5)$} & $0.06140$ & $0.03812$ & $0.02180$ & $0.00931$ & $0.02177$ & $0.20754$\\
{Enhanced West $(N=5)$} & $0.06139$ & $0.02979$ & $0.01802$ & $0.01556$ & $0.01806$ & $0.18922$\\
{GMRC $(N=5)$} & $\mathbf{0.02357}$ & $\mathbf{0.00017}$ & $\mathbf{0.00018}$ & $\mathbf{0.00010}$ & $\mathbf{0.00017}$ & $\mathbf{0.01820}$\\
{OGGMR  $(N=5)$} & $0.04938$ & $0.00084$ & $0.00220$ & $0.00064$ & $\mathbf{0.00017}$ & $\mathbf{0.01820}$\\
\hline
\end{tabular}}
\label{table1}
\end{table*}
\vspace*{-10pt}
\subsection{Experimental Setups} \label{subsec4.1}
\vspace{-2 pt}
In order to visualize how our algorithm works using the proposed measure, a GMR scenario presented in \cite{crouse2011look} is adopted, whose parameters are as follows:
\vspace*{-0.25cm}
\setlength{\abovedisplayskip}{10pt}
\setlength{\belowdisplayskip}{10pt}
\begin{flalign}
&\scalemath{0.95}{\boldsymbol{\pi}  = [0.03, 0.18, 0.12, 0.19, 0.02, 0.16, 0.06, 0.1, 0.08, 0.06]^{T}} \nonumber \\
&\scalemath{0.95}{\boldsymbol{\mu_{P}}  = [1.45, 2.20, 0.67, 0.48, 1.49, 0.91, 1.01, 1.42, 2.77, 0.89]^{T}} \nonumber \\
&\scalemath{0.95}{\boldsymbol{\Sigma_{P}}  = [0.0487, 0.0305, 0.1171, 0.0174, 0.0295, 0.0102, 0.0323,} \nonumber \\
&\hspace{5cm}\scalemath{0.95}{0.0380, 0.0115, 0.0679]^{T}}. \nonumber \end{flalign}
This GMR problem is a one-dimensional MoG with ten components ($M = 10$) that can be reduced to $N=1, 3$, and $5$ components. Various GMR algorithms, including GMRC \cite{schieferdecker2009gaussian} (with 500 number of samples), West \cite{west1993approximating}, Enhanced West \cite{chang2010scalable}, and OGGMR, have been used with different measurements, such as KL \cite{ardeshiri2012mixture}, ISE \cite{d2021likeness}, NISE \cite{d2021likeness}, TSL \cite{liu2012shape}, CS \cite{kampa2011closed}, and our metric (Table \ref{table1}). Experiments were performed using Python on the NVIDIA Tesla V100-SXM2-16GB. 
\vspace{-10pt}
\subsection{Evaluation Metrics} \label{subsec4.2}
\vspace{-3pt}
This section presents evaluation metrics to illustrate the effectiveness and efficiency of the proposed distance and algorithm. In order to examine the performance and complexity of the GMR algorithms for each measurement, we will use loss function values (objective value) and execution time, respectively. The results for run-time are averaged over $500$ runs.


\vspace{-10pt}
\subsection{Experimental Results} \label{subsec4.3}
\begin{figure} 
    \centering
    \includegraphics[width= 0.8\linewidth]{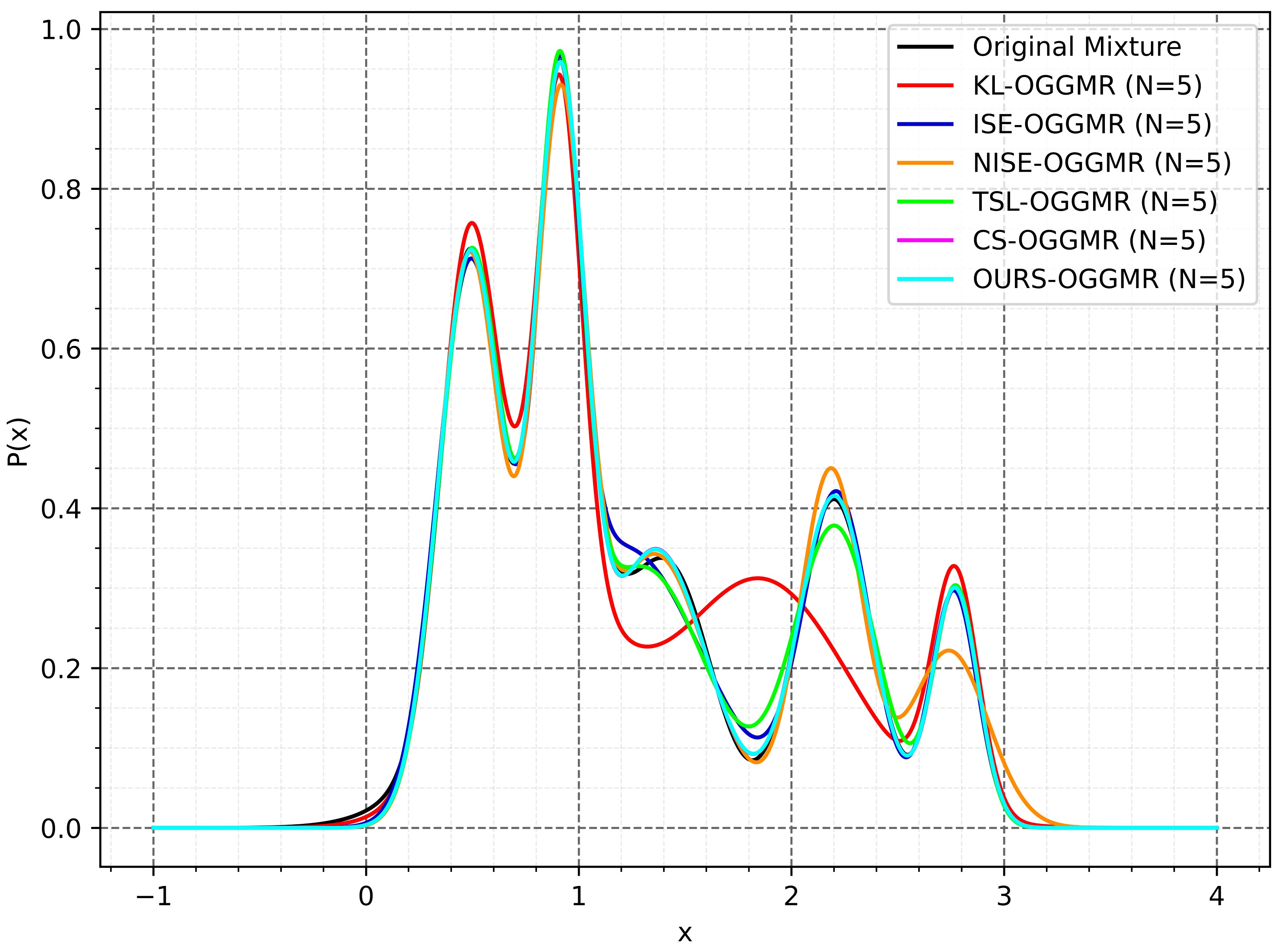}
    \caption{A comparison of different measures in the OGGMR algorithm for reducing $10$ components to $5$ components.}
    \vspace{-10pt}
    \label{fig1}
\end{figure}

\vspace{-3pt}
Table \ref{table1} displays the performance of the GMR algorithms for each measurement. The OGGMR and GMRC algorithms outperform the others for fixed dissimilarity metrics (KL, ISE, $\cdots$, OURS) and setting ($N = 1,$ $3,$ or $5$) as they combine locally shared information and remove redundant elements accurately. However, the OGGMR algorithm is significantly faster than GMRC due to its closed-form expression for MoGs, as seen in Table \ref{table2}. To ensure a fair comparison between dissimilarities, we evaluated their ability to preserve the geometric information of the original MOG in Fig. \ref{fig1}. Specifically, we compared the reduced mixtures ($N = 5$) generated by the OGGMR algorithm across different measurements that were applied to the refinement step. Using only half of the Gaussian components, the proposed distance captures all the necessary information, including orientation and position, by combining angular and magnitude distances. It preserves the geometric shape of the original MoG better than all other statistical measures, while other measurements tend to be inclusive and overlook some peaks. These results could lead to tracking loss in applications like MAP estimation, as several hypotheses may be ignored over time.
\vspace*{-23pt}
\begin{table}[htb]
\centering
\caption{Runtime of the GMRC and OGGMR algorithms in our experiment. The results are averaged over 500 runs.}
\begin{tabular}{c  c}
\hline
{\textbf{Algorithm}} & {\textbf{Execution Time (sec)}} \\
\hline
{GMRC $(N=5)$} & $4.8456 \pm 0. 2458$\\
{OGGMR $(N=5)$} & $\mathbf{0.0614 \pm 0.0138}$\\
\hline
\end{tabular}
\label{table2}
\vspace{-.74cm}
\end{table}
\section{Conclusion}
\label{sec4}
\vspace{-5pt}

This paper proposes a new probabilistic metric between two PDFs that provides a closed-form expression when MoGs are used. To demonstrate the efficiency of our metric, we developed OGGMR, an algorithm that uses the metric to solve the GMR problem. It has been found that OGGMR outperforms the state-of-the-art GMR algorithms in terms of their complexity and performance. Future work could involve extending the metric to complex-valued MoGs or evaluating it in different applications, including multi-model estimators, detection problems, and multi-label knowledge distillation tasks.
\newcommand{\BIBdecl}{\setlength{\itemsep}{0em}}
\bibliographystyle{IEEEtran}
\bibliography{refs}

\end{document}